\documentclass[11pt]{article}

\usepackage[final]{acl}

\usepackage{times}
\usepackage{latexsym}

\usepackage[T1]{fontenc}

\usepackage[utf8]{inputenc}

\usepackage{microtype}
\usepackage{amsmath}
\usepackage[most]{tcolorbox}
\usepackage{amsfonts}
\usepackage{booktabs}
\usepackage{multirow}
\usepackage{algorithm}
\usepackage{algpseudocode}
\usepackage{amsthm}

\usepackage{adjustbox}
\usepackage{inconsolata}

\usepackage{graphicx}

\title{Continual Safety Alignment via Gradient-Based Sample Selection}

\author{
Thong Bach$^{1}$\thanks{\ t.bach@deakin.edu.au} \quad
Dung Nguyen$^{1}$ \quad
Thao Minh Le$^{2}$ \quad
Truyen Tran$^{1}$
\\[0.5em]
$^{1}$ Applied Artificial Intelligence Initiative (A2I2), Deakin University \quad
$^{2}$ Pennsylvania State University
}

\begin{document}
\maketitle


\begin{abstract}
Large language models require continuous adaptation to new tasks while preserving safety alignment. However, fine-tuning on even benign data often compromises safety behaviors, including refusal of harmful requests, truthfulness, and commonsense reasoning. We investigate which training samples cause alignment drift through a data-centric lens. Our empirical analysis shows samples contribute unequally: high-gradient samples cause greater safety degradation and drive models toward pretrained distributions, while moderate-gradient samples enable task learning with minimal alignment loss. We propose gradient-based sample selection that filters high-gradient samples during fine-tuning. Across multiple model families on continual domain tasks, our method substantially improves alignment preservation while maintaining competitive task performance, without requiring curated safe data or architectural modifications. Our method is robust across selection ratios, task orderings, and diverse attack benchmarks.
\end{abstract}

\section{Introduction}

Large language models deployed in real-world applications require continuous adaptation to new domains, tasks, and evolving requirements. While initial alignment through reinforcement learning from human feedback (RLHF)~\cite{ouyang2022training}, direct preference optimization (DPO)~\cite{rafailov2024direct}, and constitutional AI~\cite{bai2022constitutional} establishes safety properties, subsequent fine-tuning often compromises these carefully cultivated behaviors~\cite{qi2024fine,yang2023shadow}.

This vulnerability presents a fundamental challenge for LLM deployment. Organizations need to customize models for specific use cases, incorporate new knowledge, and adapt to changing requirements, yet each fine-tuning step risks degrading the alignment properties that make these models safe to deploy. Even fine-tuning on benign, non-malicious datasets can unintentionally weaken safety mechanisms~\cite{qi2024fine,he2024whats}, suggesting that alignment degradation is not merely a consequence of adversarial data but a structural property of fine-tuning itself. While the data content is typically benign, the parameter updates they induce can be destructive to the alignment priors.

Continual learning research has made significant progress on retaining task performance through parameter regularization~\cite{kirkpatrick2017overcoming} and experience replay~\cite{rolnick2019experience}. However, these methods focus on preventing catastrophic forgetting of learned tasks rather than preserving alignment properties. We address a distinct problem:

\begin{tcolorbox}[colback=gray!5!white, colframe=gray!75!black, title=The Continual Safety Alignment Problem]
How can we continuously adapt LLMs to new tasks while preserving alignment properties (safety, truthfulness, helpfulness) without requiring curated safe data at each adaptation step?
\end{tcolorbox}

\begin{figure*}[t]
\centering
\includegraphics[width=\textwidth]{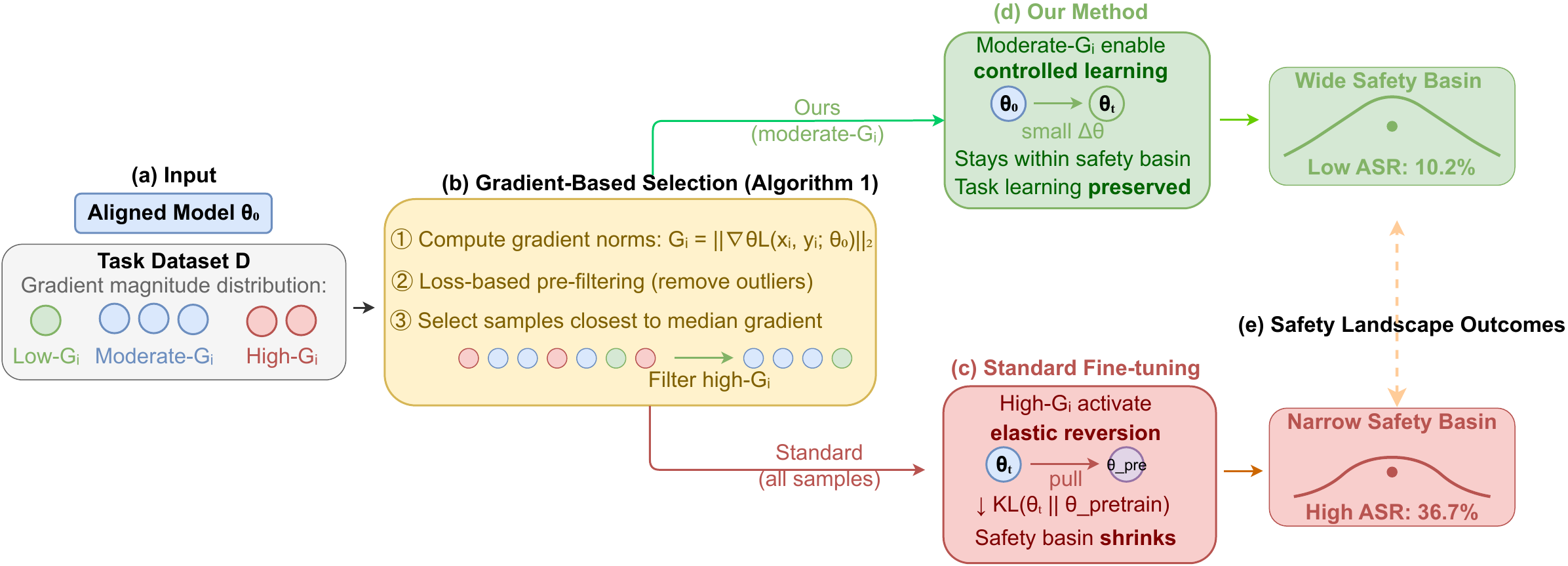}
\caption{\textbf{Overview of gradient-based sample selection for continual safety alignment.} (a) Aligned model $\theta_0$ and task dataset with varying gradient magnitudes. (b) Our method filters high-$G_i$ samples (red) and selects moderate-$G_i$ samples (blue). (c) Standard fine-tuning activates elastic reversion, pulling models toward pretrained distributions. (d) Our method enables controlled updates preserving alignment. (e) Resulting safety landscapes: narrow basin (high ASR) vs.\ wide basin (low ASR). High-gradient samples may represent alignment tension points where training tends to reverse safety modifications.}
\label{fig:overview}
\end{figure*}

Rather than constraining model architectures or requiring curated safe datasets at each adaptation step, we investigate this problem through a data-centric lens: \emph{Which training samples cause alignment drift, and can we simply avoid them?}

Recent work reveals two critical properties of aligned models that inform our approach. First, \citep{ji2024language} shows that LLMs exhibit \emph{elasticity}: a tendency to revert toward pretrained distributions during fine-tuning because the massive pretraining corpora exerts stronger influence than smaller alignment datasets. This reversion inherently degrades alignment since these pre-training corpora usually lack safety constraints. Second, \citep{peng2024navigating} discovers that aligned models occupy a ``safety basin'' in parameter space with sharp boundaries where safety collapses abruptly.

While these frameworks explain \emph{why} alignment is fragile, they do not predict \emph{which samples} cause drift or \emph{how} to maintain alignment during task adaptation. We hypothesize that training samples activate elastic reversion unequally: samples where aligned predictions diverge substantially from task targets (high gradients) may reverse alignment modifications, while moderate-gradient samples enable learning with minimal drift.

We validate this hypothesis through systematic experiments and propose a practical data-centric solution. Our contributions are:

\textbf{Empirical finding.} We provide empirical evidence that per-sample gradient magnitude predicts safety drift in continual fine-tuning. Through KL-divergence analysis, we show high-gradient samples shift models toward pretrained distributions (Table~\ref{tab:kl}), and demonstrate that standard remedies such as gradient clipping are insufficient (Section~\ref{sec:clipping}), establishing that the issue is sample-specific rather than purely magnitude-based.

\textbf{Mechanistic analysis.} We characterize high-gradient samples as format mismatches---short-answer tasks (classification, closed QA) where the aligned model's verbose output distribution diverges from terse targets---connecting our gradient-based findings with independent observations from representation-based data selection~\cite{hsiung2025llm} and benign data auditing~\cite{he2024whats}.

\textbf{Practical recipe.} We propose gradient-based sample selection with a tunable safety-task tradeoff, validated across three model families, multiple task sequences and orderings, and diverse safety benchmarks including AdvBench and HarmBench. Our method requires no curated safe data or architectural modifications.

\section{Background}
\label{sec:background}
\subsection{Measuring Alignment via Safety Basins}

To study alignment dynamics quantitatively, we adopt the safety basin framework from \citep{peng2024navigating}. This framework conceptualizes alignment as occupying a region in parameter space rather than a binary property.

Given aligned parameters $\theta_{\text{align}}$, the safety landscape is defined by perturbing along direction $\hat{d}$: $f(\alpha) = S(\theta_{\text{align}} + \alpha\hat{d})$ where $S(\cdot)$ measures attack success rate. This defines a \emph{safety basin} $\mathcal{B} = \{\theta : S(\theta) \leq S_{\text{threshold}}, \theta \text{ connected to } \theta_{\text{align}}\}$, the connected region in parameter space where safety properties hold.

A critical empirical finding is that safety basins have \emph{sharp boundaries}: safety exhibits step-function collapse when crossing the boundary, with minimal graceful degradation. This geometry makes large parameter updates particularly dangerous. A single large step can push models from safe to unsafe, while many small updates might stay within the basin.

The VISAGE score quantifies basin volume by averaging safety margin across random perturbation directions:
\begin{equation}
\text{VISAGE} = \mathbb{E}_{\alpha \sim U(-a,a)}[S_{\max} - S(\alpha)] \quad \text{s.t. } S < S_{\max},
\label{eq:visage}
\end{equation}
where $S_{\text{max}} = 100\%$ represents complete safety failure. 
Higher VISAGE indicates larger safety basins and more robust alignment. We define alignment drift at step $t$ as: $\Delta_{\text{align}}(t) = \text{VISAGE}(\theta_0) - \text{VISAGE}(\theta_t)$. In our experiments, we compute VISAGE using $N=100$ random perturbation directions, with perturbation range $a$ calibrated per model. The safety margin is averaged across all directions, providing robust measurement beyond any single perturbation slice. See Appendix~\ref{app:background} for extended discussion.

\subsection{Alignment Fragility and Elasticity}
Alignment degradation during fine-tuning reflects structural properties of the learning process rather than just adversarial data. Previous work \citep{qi2024fine} shows that as few as 10 examples can compromise safety, while \citep{he2024whats} demonstrates that even benign datasets weaken safety mechanisms.

The elasticity framework~\citep{ji2024language} provides theoretical grounding for this phenomenon: language models resist alignment modifications and rebound toward pretrained behavior under perturbation. The elastic force is proportional to dataset size: $F_{\text{elastic}} \propto |\mathcal{D}_i| \cdot \Delta D_{\text{KL}}(p_\theta \| p_{\mathcal{D}_i})$. Since pretrain corpora ($|\mathcal{D}_p|$) vastly exceed alignment datasets ($|\mathcal{D}_a|$), the pretrained distribution exerts orders of magnitude stronger ``pull'' on model behavior.

This asymmetry predicts two phenomena: \emph{resistance} (pretrained models resist initial alignment) and \emph{rebound} (aligned models revert toward pretrained behavior under fine-tuning). Our hypothesis extends this framework to the sample level: training samples where aligned predictions diverge substantially from task targets, indicated by high gradient magnitudes, may specifically activate this elastic reversion force, accelerating drift toward pretrained distributions and out of safety basins.

\subsection{Problem Formulation}

We formalize the continual safety alignment problem, distinguishing it from standard continual learning.

\paragraph{Setting.} Consider an aligned model $\theta_0$ that must learn $T$ tasks sequentially from datasets $\{\mathcal{D}_1, \mathcal{D}_2, \ldots, \mathcal{D}_T\}$. Each dataset $\mathcal{D}_t = \{(x_i, y_i)\}_{i=1}^{N_t}$ contains input-output pairs for task $t$.

\paragraph{Standard Continual Learning.} Traditional continual learning minimizes task loss while preventing forgetting of previous tasks:
\begin{equation}
    \theta_t
    = \arg\min_{\theta}
    \mathcal{L}_t(\theta; \mathcal{D}_t)
    + \lambda \cdot \text{Reg}(\theta; \theta_{t-1})
\end{equation}
where $\text{Reg}(\cdot)$ penalizes deviation from previous parameters (e.g., EWC~\cite{kirkpatrick2017overcoming}) or replays stored examples~\cite{rolnick2019experience}.

\paragraph{Continual Safety Alignment.} We impose an additional constraint, alignment preservation:
\begin{equation}
    \theta_t = \arg\min_{\theta} \mathcal{L}_t(\theta; \mathcal{D}_t) \quad \text{s.t.} \quad \theta_t \in \mathcal{B}
    \label{eq:csa_constraint}
\end{equation}
where $\mathcal{B}$ is the safety basin. The model must remain within the basin throughout training, not just at convergence. Since pretrained models lie outside the safety basin and elasticity pulls parameters toward pretrained configurations, fine-tuning must resist this force to preserve alignment.

\paragraph{Alignment Drift Metric.} We quantify alignment degradation using VISAGE (Eq.~\ref{eq:visage}). Define alignment drift at step $t$ as:
\begin{equation}
    \Delta_{\text{align}}(t) = \text{VISAGE}(\theta_0) - \text{VISAGE}(\theta_t)
    \label{eq:drift}
\end{equation}

The continual safety alignment objective becomes:
\begin{equation}
    \min_{\theta_1, \ldots, \theta_T} \sum_{t=1}^{T} \mathcal{L}_t(\theta_t; \mathcal{D}_t) \quad \text{s.t.} \quad \Delta_{\text{align}}(t) \leq \delta \quad \forall t
    \label{eq:csa_objective}
\end{equation}
where $\delta$ is a tolerance threshold for acceptable alignment drift.

\paragraph{Challenges.} The constraint in Eq.~\ref{eq:csa_objective} is difficult to enforce directly: \textbf{1. Measurement cost}: Computing VISAGE requires evaluating safety across multiple perturbation directions, which is expensive during training. \textbf{2. Non-differentiability}: VISAGE is not differentiable with respect to $\theta$, precluding gradient-based constraint enforcement. \textbf{3. Sample-level opacity}: The constraint operates at the model level, but training operates at the sample level, so it is unclear which samples contribute to drift.

\paragraph{Toward a data-centric solution.} The third challenge, sample-level opacity, motivates our approach. Rather than enforcing model-level constraints (expensive and non-differentiable), we ask: \emph{which training samples accelerate alignment drift?} If samples contribute unequally, we can filter those causing disproportionate drift. This shifts the problem from constrained optimization to sample selection.

\section{Analysis: Which Samples Cause Drift?}
\label{sec:analysis}
\begin{figure}[t]
    \centering
    \includegraphics[width=\columnwidth]{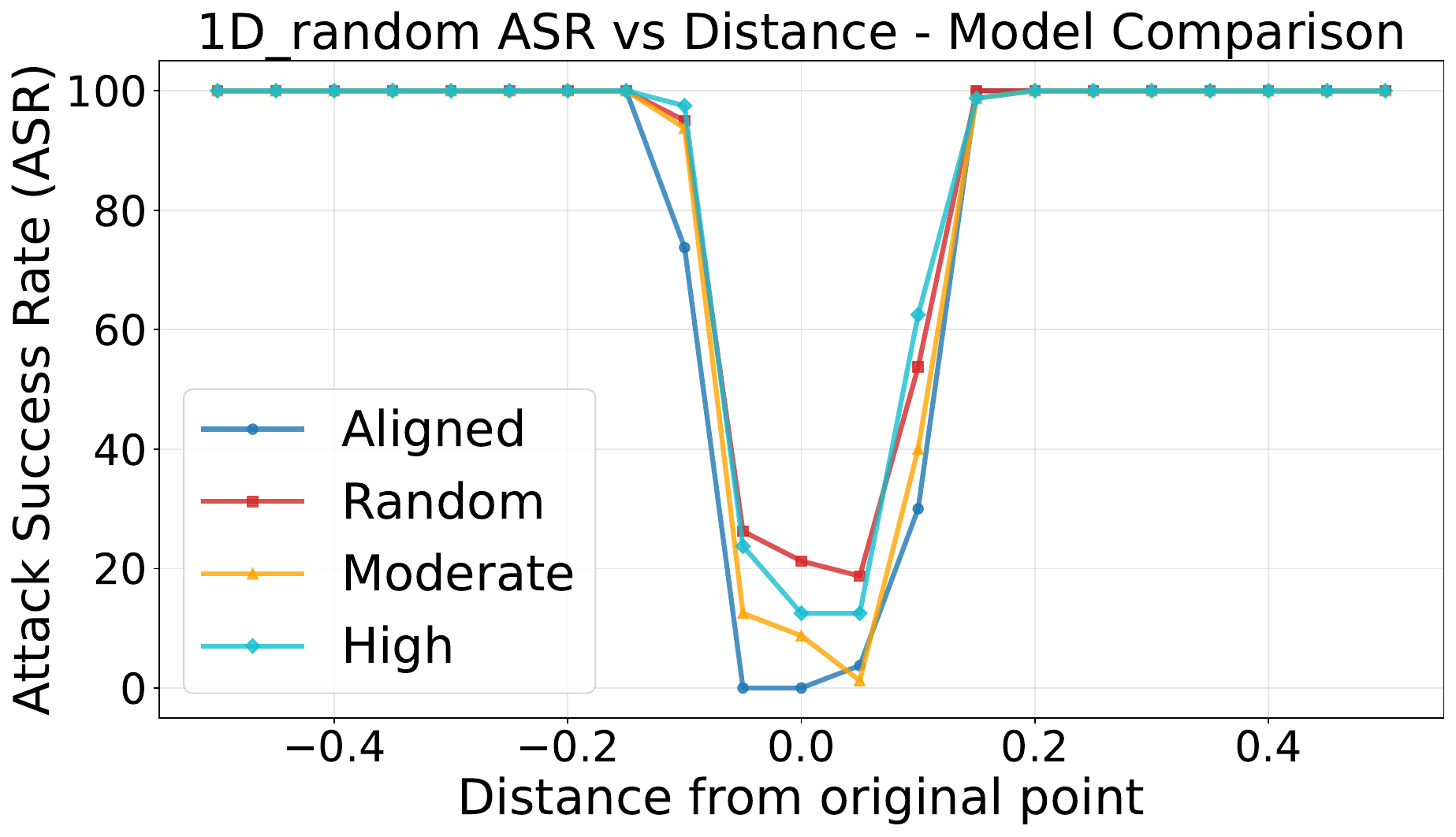}
    \caption{Safety landscape visualization for Qwen-2.5-7B-Instruct after fine-tuning on Dolly. Each curve shows the attack success rate under parameter perturbations. High-$G_i$ selection (cyan) creates a narrow basin where small perturbations cause safety collapse. Moderate-$G_i$ (orange) maintains a wide basin comparable to the original model (blue), demonstrating robust alignment preservation.}

    \label{fig:basin}
    \end{figure}
Following the data-centric perspective introduced in Section~\ref{sec:background}, we investigate: \emph{do all samples contribute equally to alignment drift, or can sample-level properties predict drift risk?} 

\textbf{Hypothesis.} We hypothesize that per-sample gradient magnitude indicates drift risk. High-gradient samples occur where aligned predictions diverge substantially from task targets, precisely where alignment training modified behavior away from pretrained tendencies. Training on these samples may reverse those modifications, activating elastic reversion.

\textbf{Experimental design.} We fine-tune LLaMA-3.1-8B-Instruct~\cite{llama3} and Qwen-2.5-7B-Instruct~\cite{qwen2.5} on Dolly~\cite{DatabricksBlog2023DollyV2} (15K benign instruction-following examples). We compare three selection strategies, each using 3,000 samples (20\%): (1) \textbf{Random} (baseline), (2) \textbf{High-$G_i$} (top 20\% by gradient norm $G_i = \|\nabla_\theta \mathcal{L}(x_i, y_i; \theta_0)\|_2$), and (3) \textbf{Moderate-$G_i$} (20\% closest to median gradient norm). All use identical hyperparameters (lr=$2 \times 10^{-5}$, 3 epochs, AdamW). We evaluate VISAGE~\cite{peng2024navigating}, attack success rate (ASR) on AdvBench~\cite{zou2023universal}, and task performance. See Appendix~\ref{app:setup} for full details.

\paragraph{Why not low-gradient samples?} To characterize the full gradient-safety spectrum, we evaluate Low-$G_i$ selection (bottom 20\% by gradient norm) as a complete baseline across all three model families (Table~\ref{tab:low_gi}).

\begin{table}[t]
\begin{adjustbox}{width=0.48\textwidth}
\centering\small
\begin{tabular}{llccc}
\toprule
\textbf{Model} & \textbf{Selection} & \textbf{ASR}$\downarrow$ & \textbf{TruthfulQA} & \textbf{Task Avg} \\
\midrule
\multirow{2}{*}{Qwen2.5-7B} & Low-$G_i$ & \textbf{8.4} & \textbf{50.2} & 60.1 \\
& Moderate-$G_i$ & 10.2 & 42.5 & \textbf{60.9} \\
\midrule
\multirow{2}{*}{LLaMA-3.1-8B} & Low-$G_i$ & \textbf{13.3} & \textbf{45.5} & 44.5 \\
& Moderate-$G_i$ & 18.3 & 41.5 & \textbf{46.4} \\
\midrule
\multirow{2}{*}{Qwen3-4B} & Low-$G_i$ & \textbf{3.0} & \textbf{48.6} & 56.3 \\
& Moderate-$G_i$ & 6.0 & 42.8 & \textbf{58.1} \\
\bottomrule
\end{tabular}
\end{adjustbox}
\caption{Low-$G_i$ vs.\ Moderate-$G_i$ selection. Low-$G_i$ provides better safety preservation but trades 0.8--1.9 points of task performance, revealing a Pareto tradeoff along the gradient spectrum.}
\label{tab:low_gi}
\end{table}

Low-$G_i$ provides better safety preservation across all models but consistently trades 0.8--1.9 points of task performance. This reveals a clean Pareto tradeoff: Low-$G_i$ for best safety, Moderate-$G_i$ for best task performance, High-$G_i$ worst on both dimensions. This strengthens our core finding that gradient magnitude monotonically predicts alignment drift. We recommend Moderate-$G_i$ as a practical default; safety-critical applications may prefer Low-$G_i$.

\subsection{Results Support the Hypothesis}

\begin{table}[t]
    \centering
    \begin{adjustbox}{width=0.48\textwidth}
    \begin{tabular}{llccc}
    \toprule
    \textbf{Model} & \textbf{Selection} & \textbf{VISAGE} & \textbf{Retain} & \textbf{ASR}$\downarrow$ \\
    \midrule
    \multirow{4}{*}{Qwen-2.5-7B} 
    & Aligned & 78.5 & - & 2.1 \\
    & High-$G_i$ & 48.8 & 62.2\% & 18.4 \\
    & Random & 57.0 & 72.6\% & 12.7 \\
    & Moderate-$G_i$ & \textbf{65.5} & \textbf{83.4\%} & \textbf{5.8} \\
    \midrule
    \multirow{4}{*}{LLaMA-3.1-8B}
    & Aligned & 67.7 & - & 3.2 \\
    & High-$G_i$ & 48.9 & 72.2\% & 15.1 \\
    & Random & 52.8 & 78.0\% & 9.8 \\
    & Moderate-$G_i$ & \textbf{59.3} & \textbf{87.6\%} & \textbf{4.9} \\
    \bottomrule
    \end{tabular}
    \end{adjustbox}
    \caption{Alignment preservation across selection strategies. High-gradient selection causes the largest degradation (62-72\% retention); moderate-gradient selection preserves 83-88\% with the lowest ASR. Results shown for two model families.}
    \label{tab:visage}
    \end{table}
\paragraph{High-gradient samples show greater alignment drift} Table~\ref{tab:visage} shows high-$G_i$ selection retains only 62-72\% of original alignment and increases ASR by 5-9$\times$, while moderate-$G_i$ selection preserves 83-88\% with only 1.5-2$\times$ ASR increase. Figure~\ref{fig:basin} visualizes this: high-$G_i$ dramatically narrows the safety basin, while moderate-$G_i$ preserves basin width comparable to the original model.

\paragraph{Evidence for elastic reversion mechanism.} 
\begin{table}[t]
    \centering\small
    \begin{tabular}{llcc}
    \toprule
    \textbf{Model} & \textbf{Selection} & \textbf{KL}$_\text{pretrain}$ & \textbf{KL}$_\text{aligned}$ \\
    \midrule
    \multirow{3}{*}{Qwen-2.5-7B}
    & High-$G_i$ & 0.14 & 0.25 \\
    & Random & 0.15 & 0.23 \\
    & Moderate-$G_i$ & \textbf{0.18} & \textbf{0.17} \\
    \midrule
    \multirow{3}{*}{LLaMA-3.1-8B}
    & High-$G_i$ & 0.51 & 0.14 \\
    & Random & 0.55 & 0.11 \\
    & Moderate-$G_i$ & \textbf{0.68} & \textbf{0.06} \\    
    \bottomrule
    \end{tabular}
    \caption{KL-divergence to pretrained and aligned distributions of different sample selection methods. High-$G_i$ samples are associated with shifts toward pretrained distributions (lower KL$_\text{pretrain}$) and away from aligned distributions (higher KL$_\text{aligned}$).}
    \label{tab:kl}
    \end{table}

Table~\ref{tab:kl} provides evidence consistent with the elastic reversion hypothesis. Training on High-$G_i$ samples correlates with movement toward pretrained distributions (lower KL$_\text{pretrain}$) and \emph{away from} aligned distributions (higher KL$_\text{aligned}$). This pattern is consistent with high-gradient samples activating the elastic reversion force described by \citep{ji2024language}: they represent alignment tension points where training reverses safety modifications.

\subsection{Gradient Direction Analysis: Preliminary Investigation}
\label{sec:direction_analysis}

To explore whether high-gradient samples have gradients aligned with the reversion direction $\mathbf{r} = \theta_{\text{pretrain}} - \theta_{\text{aligned}}$, we analyze gradient directions across parameter subsets. Direct cosine similarity in billion-dimensional parameter spaces yields near-zero values due to concentration of measure. We address this by computing TopK-Cosine: cosine similarity restricted to the $k$ dimensions where alignment training induced the largest parameter changes (see Appendix~\ref{appendix:gradient_direction} for methodology details).

Table~\ref{tab:direction_analysis} presents results for Qwen2.5-7B-Instruct and LLaMA-3.1-8B-Instruct. High-gradient samples exhibit higher directional alignment with the reversion direction compared to moderate-gradient samples in final-layer parameters, though the specific components vary by architecture: V/O projections in Qwen2.5 (TopK-Cosine 0.119 vs 0.104 for V, $r=0.41$) and MLP layers in LLaMA (0.104 vs 0.102, $r=0.18$). Besides, middle layers show no directional effect in either model ($|r|<0.07$, $p>0.3$), confirming that the signal localizes to alignment-critical parameters in the final transformer layer.

\begin{table}[t]
\centering

\begin{adjustbox}{width=0.48\textwidth}
\begin{tabular}{llcccc}
\toprule
\textbf{Model} & \textbf{Parameter} & \textbf{HIGH} & \textbf{MOD} & $\mathbf{r}$ & $\mathbf{p}$ \\
\midrule
\multirow{3}{*}{Qwen2.5-7B} 
& Last\_V & 0.119 & 0.104 & 0.41 & $<10^{-3}$ \\
& Last\_O & 0.276 & 0.244 & 0.39 & $<10^{-3}$ \\
& Middle & $-$0.004 & $-$0.004 & 0.06 & 0.38 \\
\midrule
\multirow{3}{*}{LLaMA-3.1-8B} 
& Last\_MLP & 0.104 & 0.102 & 0.18 & $<0.01$ \\
& Last\_V & $-$0.029 & $-$0.033 & 0.33 & $<10^{-3}$ \\
& Middle & $-$0.020 & $-$0.024 & 0.03 & 0.72 \\
\bottomrule
\end{tabular}
\end{adjustbox}
\caption{TopK-Cosine ($k$=1000) between gradients and reversion direction. HIGH and MOD denote the top 20\% and middle 20\% by gradient norm. Directional alignment appears in final-layer parameters with no effect in middle layers.}
\label{tab:direction_analysis}
\end{table}

These findings provide preliminary support for the elastic reversion hypothesis: \emph{high-gradient samples may produce gradients that partially reverse final-layer modifications learned during alignment.} However, the modest correlation strengths ($r = 0.18$--$0.41$) and 
architectural variation (attention outputs in Qwen2.5 vs. MLP in LLaMA) suggest this relationship is complex. We note that our selection method relies only on gradient magnitude, not direction, and is effective regardless of whether the directional hypothesis fully holds. We emphasize the gradient clipping comparison (Section~\ref{sec:clipping}) and empirical stability as the primary support for why accumulation of small gradient steps does not undermine our method.

\subsection{Implications for Method Design}

These findings validate our data-centric approach: \emph{filter high-gradient samples during fine-tuning}. Moderate-gradient samples provide sufficient learning signal for task adaptation while avoiding alignment-reversing effects. This motivates our gradient-based selection method (Section~\ref{sec:method}), which requires no curated safe data, only gradient computation on the task dataset itself.

\section{Method: Gradient-Based Sample Selection}
\label{sec:method}

Our analysis reveals that moderate-gradient samples enable task learning with minimal alignment drift. We operationalize this insight through a batch selection algorithm that filters high-gradient samples during fine-tuning.

\subsection{Algorithm}

Algorithm~\ref{alg:gradient_selection} operates in three stages: (1) loss-based pre-filtering removes extreme samples (very low loss = memorized; very high loss = outliers), retaining ${\sim}68\%$ of candidates, (2) gradient computation on filtered candidates, reducing computational cost by not computing gradients for all candidates, and (3) median-based selection chooses samples closest to median gradient norm $\mu_G$, avoiding both high-gradient (causing alignment drift) and low-gradient samples (providing minimal learning signal).

\begin{algorithm}
\caption{Gradient-Based Sample Selection}
\label{alg:gradient_selection}
\begin{algorithmic}[1]
\Require Batch $\mathcal{B}$, model $\theta$, selection ratio $\rho = 0.2$
\Ensure Selected samples $\mathcal{S}$
\State Compute losses: $L_i = \mathcal{L}(x_i, y_i; \theta)$ for all $(x_i, y_i) \in \mathcal{B}$
\State Filter: $\mathcal{C} \leftarrow \{(x_i, y_i) : L_i \in [\mu_L - \sigma_L, \mu_L + \sigma_L]\}$
\State Compute gradient norms: $G_i = \|\nabla_\theta \mathcal{L}(x_i, y_i; \theta)\|_2$ for $(x_i, y_i) \in \mathcal{C}$
\State $\mu_G \leftarrow \text{median}(\{G_i\})$
\State Select $\lfloor \rho |\mathcal{B}| \rfloor$ samples closest to $\mu_G$ as $\mathcal{S}$
\State \Return $\mathcal{S}$
\end{algorithmic}
\end{algorithm}

\textbf{Key design choices.} We use median (not mean) for robustness against heavy-tailed gradient distributions. Selection ratio $\rho \in [0.15, 0.25]$ balances quality vs. cost; we use $\rho = 0.2$. Pre-filtering reduces gradient computation by ${\sim}32\%$.

\subsection{Sensitivity to Selection Ratio $\rho$}
\label{sec:sensitivity}
We conduct a systematic sensitivity analysis on $\rho \in \{0.1, 0.2, 0.4, 0.6\}$ using Qwen3-4B across the full 4-task continual learning pipeline (Table~\ref{tab:sensitivity}).

\begin{table}[t]
\centering\small
\begin{tabular}{ccccc}
\toprule
$\rho$ & \textbf{ASR}$\downarrow$ & \textbf{TruthfulQA} & \textbf{HellaSwag} & \textbf{Task Avg} \\
\midrule
0.1 & 2.7 & 43.3 & 69.7 & 59.3 \\
\textbf{0.2} & \textbf{6.0} & \textbf{42.8} & \textbf{70.1} & \textbf{58.1} \\
0.4 & 5.5 & 43.9 & 69.3 & 57.1 \\
0.6 & 6.6 & 42.2 & 69.5 & 57.2 \\
\bottomrule
\end{tabular}
\caption{Sensitivity to selection ratio $\rho$ on Qwen3-4B. ASR remains consistently low across $\rho \in [0.1, 0.4]$ (2.7--6.0\%), all substantially better than baseline (16.6\%) and random (11.8\%).}
\label{tab:sensitivity}
\end{table}

Results are robust across $\rho \in [0.1, 0.4]$: ASR remains consistently low (2.7--6.0\%), all substantially better than baseline (16.6\%) and random sampling (11.8\%). Smaller $\rho$ (stricter filtering) provides slightly better safety (2.7\% at $\rho=0.1$) at marginal task performance cost (59.3\% vs 58.1\%). Larger $\rho$ (less filtering) converges toward random sampling behavior. We recommend $\rho = 0.2$ as a default, but practitioners can tune based on safety-task priorities.

\section{Experiments}
\label{sec:experiments}

We evaluate our gradient-based sample selection method on continual safety alignment across multiple model families and diverse task sequences. Our experiments demonstrate that filtering high-gradient samples better preserves both task performance and safety alignment throughout sequential fine-tuning.

\subsection{Experimental Setup}
\begin{table*}[h]
\centering
\begin{adjustbox}{width=0.85\textwidth}
\begin{tabular}{llccccccc}
\toprule
\textbf{Model} & \textbf{Method} & \textbf{ASR}$\downarrow$ & \textbf{TruthfulQA} & \textbf{ARC-C} & \textbf{BoolQ} & \textbf{HellaSwag} & \textbf{Winogrande} \\
\midrule
\multirow{7}{*}{Qwen2.5}
& Baseline & 36.7 {\scriptsize$\pm$ 13.6} & 38.2 {\scriptsize$\pm$ 1.2} & 58.0 {\scriptsize$\pm$ 1.5} & \textbf{86.4} {\scriptsize$\pm$ 1.0} & 79.2 {\scriptsize$\pm$ 0.5} & \textbf{72.3} {\scriptsize$\pm$ 0.3} \\
& Random & 31.1 {\scriptsize$\pm$ 14.3} & 38.4 {\scriptsize$\pm$ 1.1} & 58.1 {\scriptsize$\pm$ 0.6} & 85.9 {\scriptsize$\pm$ 2.0} & 79.4 {\scriptsize$\pm$ 0.3} & \textbf{72.3} {\scriptsize$\pm$ 0.5} \\
& KL & 33.5 {\scriptsize$\pm$ 13.4} & 37.8 {\scriptsize$\pm$ 1.3} & 58.3 {\scriptsize$\pm$ 1.7} & 86.2 {\scriptsize$\pm$ 1.4} & 79.1 {\scriptsize$\pm$ 0.4} & \textbf{72.3} {\scriptsize$\pm$ 0.4} \\
& O-LoRA & 16.5 {\scriptsize$\pm$ 19.7}& \textbf{42.9} {\scriptsize$\pm$ 1.0} & 56.6 {\scriptsize$\pm$ 0.9} & 86.1 {\scriptsize$\pm$ 1.2} & 79.3 {\scriptsize$\pm$ 0.4} & 71.8 {\scriptsize$\pm$ 0.6} \\
& EWC & 17.4 {\scriptsize$\pm$ 6.7} & 38.1 {\scriptsize$\pm$ 0.3} & 57.7 {\scriptsize$\pm$ 0.4} & 86.8 {\scriptsize$\pm$ 0.2} & 79.2 {\scriptsize$\pm$ 0.1} & 71.7 {\scriptsize$\pm$ 0.3} \\
& Grad.\ Clip & 31.2 {\scriptsize$\pm$ 13.8} & 38.2 {\scriptsize$\pm$ 1.1} & 57.9 {\scriptsize$\pm$ 1.4} & 86.3 {\scriptsize$\pm$ 1.1} & 79.3 {\scriptsize$\pm$ 0.4} & 72.1 {\scriptsize$\pm$ 0.4} \\
& Moderate-$G_i$ & \textbf{10.2} {\scriptsize$\pm$ 7.1} & 42.5 {\scriptsize$\pm$ 1.3} & \textbf{59.6} {\scriptsize$\pm$ 1.7} & 86.3 {\scriptsize$\pm$ 1.5} & \textbf{79.9} {\scriptsize$\pm$ 0.1} & 71.7 {\scriptsize$\pm$ 0.9} \\
\midrule
\multirow{7}{*}{LLaMA-3.1}
& Baseline & 44.2 {\scriptsize$\pm$ 22.5} & 37.8 {\scriptsize$\pm$ 0.7} & 56.3 {\scriptsize$\pm$ 1.3} & 84.8 {\scriptsize$\pm$ 1.0} & 77.9 {\scriptsize$\pm$ 0.5} & 73.8 {\scriptsize$\pm$ 0.5} \\
& Random & 31.9 {\scriptsize$\pm$ 23.3} & 38.6 {\scriptsize$\pm$ 1.7} & \textbf{56.9} {\scriptsize$\pm$ 1.5} & 84.8 {\scriptsize$\pm$ 0.6} & 78.0 {\scriptsize$\pm$ 0.3} & 73.6 {\scriptsize$\pm$ 0.5} \\
& KL & 43.6 {\scriptsize$\pm$ 21.6}& 38.6 {\scriptsize$\pm$ 0.8} & 56.4 {\scriptsize$\pm$ 1.5} & \textbf{84.9} {\scriptsize$\pm$ 1.0} & 78.0 {\scriptsize$\pm$ 0.6} & \textbf{74.0} {\scriptsize$\pm$ 0.5} \\
& O-LoRA & 23.3 {\scriptsize$\pm$ 28.0} & 40.0 {\scriptsize$\pm$ 1.0} & 56.4 {\scriptsize$\pm$ 1.2} & 84.7 {\scriptsize$\pm$ 0.6} & \textbf{78.3} {\scriptsize$\pm$ 0.6} & \textbf{74.0} {\scriptsize$\pm$ 0.7} \\
& EWC & 40.2 {\scriptsize$\pm$ 11.8} & 38.0 {\scriptsize$\pm$ 0.3} & 56.0 {\scriptsize$\pm$ 0.7} & 84.9 {\scriptsize$\pm$ 0.2} & 78.0 {\scriptsize$\pm$ 0.1} & 74.0 {\scriptsize$\pm$ 0.2} \\
& Moderate-$G_i$ & \textbf{18.3} {\scriptsize$\pm$ 17.3} & \textbf{41.5} {\scriptsize$\pm$ 2.6} & 56.0 {\scriptsize$\pm$ 1.5} & 84.8 {\scriptsize$\pm$ 0.6} & 77.9 {\scriptsize$\pm$ 0.7} & 73.7 {\scriptsize$\pm$ 0.5} \\
\midrule
\multirow{7}{*}{Qwen3}
& Baseline & 16.6 {\scriptsize$\pm$ 7.7} & 39.7 {\scriptsize$\pm$ 0.7} & 59.8 {\scriptsize$\pm$ 1.5} & \textbf{86.2} {\scriptsize$\pm$ 0.5} & \textbf{71.0} {\scriptsize$\pm$ 1.2} & 68.7 {\scriptsize$\pm$ 0.4} \\
& Random & 11.8 {\scriptsize$\pm$ 5.8} & 40.1 {\scriptsize$\pm$ 1.6} & \textbf{60.2} {\scriptsize$\pm$ 1.1} & 85.6 {\scriptsize$\pm$ 1.2} & 70.6 {\scriptsize$\pm$ 0.8} & 68.7 {\scriptsize$\pm$ 0.7} \\
& KL & 17.8 {\scriptsize$\pm$ 7.0} & 39.7 {\scriptsize$\pm$ 0.8} & 59.5 {\scriptsize$\pm$ 1.2} & 86.0 {\scriptsize$\pm$ 0.4} & \textbf{71.0} {\scriptsize$\pm$ 1.2} & \textbf{69.0} {\scriptsize$\pm$ 0.7} \\
& O-LoRA & 6.8 {\scriptsize$\pm$ 6.5} & 42.3 {\scriptsize$\pm$ 1.6} & 59.5 {\scriptsize$\pm$ 1.3} & 85.2 {\scriptsize$\pm$ 0.8} & 70.7 {\scriptsize$\pm$ 1.1} & 68.7 {\scriptsize$\pm$ 0.7} \\
& EWC & 14.1 {\scriptsize$\pm$ 3.1} & 40.5 {\scriptsize$\pm$ 0.2} & 59.3 {\scriptsize$\pm$ 0.3} & 85.7 {\scriptsize$\pm$ 0.1} & 71.1 {\scriptsize$\pm$ 0.1} & 68.9 {\scriptsize$\pm$ 0.3} \\
& Moderate-$G_i$ & \textbf{6.0} {\scriptsize$\pm$ 5.4} & \textbf{42.8} {\scriptsize$\pm$ 1.6} & 59.2 {\scriptsize$\pm$ 1.3} & 84.6 {\scriptsize$\pm$ 1.4} & 70.1 {\scriptsize$\pm$ 0.8} & 68.5 {\scriptsize$\pm$ 0.8} \\
\bottomrule
\end{tabular}
\end{adjustbox}
\caption{Alignment preservation metrics, checkpoint-averaged (mean $\pm$ std over three seeds). Safety via attack success rate (ASR, lower better), truthfulness via TruthfulQA, and general capabilities via ARC-Challenge, BoolQ, HellaSwag, and Winogrande.}
\label{tab:alignment}
\end{table*}

\begin{table*}[h]
\centering
\begin{adjustbox}{width=0.8\textwidth}
\begin{tabular}{llccccc}
\toprule
\textbf{Model} & \textbf{Method} & \textbf{After GSM8K} & \textbf{After MedMCQA} & \textbf{After Squad} & \textbf{Avg} \\
\midrule
\multirow{7}{*}{Qwen2.5} 
& Baseline & 58.2 {\scriptsize$\pm$ 0.2} & 57.8 {\scriptsize$\pm$ 0.8} & 61.3 {\scriptsize$\pm$ 0.9} & 59.1 \\
& Random & 57.7 {\scriptsize$\pm$ 0.7} & 56.8 {\scriptsize$\pm$ 4.1} & 62.2 {\scriptsize$\pm$ 2.6} & 58.9 \\
& KL & 58.1 {\scriptsize$\pm$ 0.2} & \textbf{63.1} {\scriptsize$\pm$ 0.3} & 61.7 {\scriptsize$\pm$ 2.1} & \textbf{61.0} \\
& O-LoRA & 54.7 {\scriptsize$\pm$ 0.6} & 59.2 {\scriptsize$\pm$ 0.5} & 53.0 {\scriptsize$\pm$ 0.8} & 55.7 \\
& EWC & 57.8 {\scriptsize$\pm$ 0.2} & 57.3 {\scriptsize$\pm$ 1.2} & 63.3 {\scriptsize$\pm$ 0.5} & 59.5 \\
& \textbf{Moderate-$G_i$} & \textbf{59.0} {\scriptsize$\pm$ 0.5} & 60.6 {\scriptsize$\pm$ 1.0} & \textbf{63.0} {\scriptsize$\pm$ 2.7} & 60.9 \\
\midrule
\multirow{7}{*}{LLaMA-3.1}
& Baseline & 44.4 {\scriptsize$\pm$ 0.5} & 43.8 {\scriptsize$\pm$ 1.0} & 51.5 {\scriptsize$\pm$ 1.0} & 46.5 \\
& Random & 44.4 {\scriptsize$\pm$ 1.3} & 44.2 {\scriptsize$\pm$ 0.9} & 51.9 {\scriptsize$\pm$ 1.9} & 46.8 \\
& KL & 44.5 {\scriptsize$\pm$ 0.7} & 43.9 {\scriptsize$\pm$ 0.6} & 50.7 {\scriptsize$\pm$ 2.1} & 46.4 \\
& O-LoRA & 43.6 {\scriptsize$\pm$ 0.9} & \textbf{45.7} {\scriptsize$\pm$ 0.6} & \textbf{54.5} {\scriptsize$\pm$ 0.9} & \textbf{47.9} \\
& EWC & 43.8 {\scriptsize$\pm$ 0.7} & 44.3 {\scriptsize$\pm$ 0.5} & 50.5 {\scriptsize$\pm$ 1.2} & 46.2 \\
& \textbf{Moderate-$G_i$} & \textbf{45.2} {\scriptsize$\pm$ 1.9} & 44.4 {\scriptsize$\pm$ 0.5} & 49.7 {\scriptsize$\pm$ 1.9} & 46.4 \\
\midrule
\multirow{7}{*}{Qwen3}
& Baseline & 60.7 {\scriptsize$\pm$ 0.2} & 51.2 {\scriptsize$\pm$ 0.6} & 50.6 {\scriptsize$\pm$ 1.0} & 54.2 \\
& Random & 60.6 {\scriptsize$\pm$ 0.9} & 54.1 {\scriptsize$\pm$ 3.0} & 50.2 {\scriptsize$\pm$ 3.9} & 55.0 \\
& KL & 59.8 {\scriptsize$\pm$ 0.9} & 52.4 {\scriptsize$\pm$ 1.8} & 52.2 {\scriptsize$\pm$ 2.4} & 54.8 \\
& O-LoRA & \textbf{60.8} {\scriptsize$\pm$ 0.3} & \textbf{59.1} {\scriptsize$\pm$ 1.0} & 56.6 {\scriptsize$\pm$ 0.4} & \textbf{58.8} \\
&EWC & 59.2 {\scriptsize$\pm$ 0.4} & 53.1 {\scriptsize$\pm$ 0.7} & 52.6 {\scriptsize$\pm$ 1.6} & 55.0 \\
& \textbf{Moderate-$G_i$} & 58.2 {\scriptsize$\pm$ 0.4} & 57.6 {\scriptsize$\pm$ 3.3} & \textbf{58.5} {\scriptsize$\pm$ 3.2} & 58.1 \\
\bottomrule
\end{tabular}
\end{adjustbox}
\caption{Checkpoint-averaged performance across all evaluation tasks at each training stage. Values represent mean $\pm$ standard deviation over three seeds. Moderate-$G_i$ maintains the highest overall performance throughout continual learning.}
\label{tab:continual_performance}
\end{table*}
\textbf{Models and tasks.} We evaluate three instruction-tuned models: Qwen2.5-7B-Instruct \citep{qwen2.5}, LLaMA-3.1-8B-Instruct \citep{llama3}, and Qwen3-4B-Instruct \citep{yang2025qwen3} (abbreviated as Qwen2.5, LLaMA-3.1, and Qwen3). Following realistic deployment scenarios, we fine-tune sequentially on four domains: Dolly \citep{DatabricksBlog2023DollyV2} (general instruction-following), GSM8K \citep{cobbe2021gsm8k} (mathematical reasoning), MedMCQA \citep{pmlr-v174-pal22a} (medical QA), and Squad\_v2 \citep{rajpurkar-etal-2018-know} (reading comprehension). Dolly is used first to reduce excessive refusal behavior in RLHF-aligned models.

\textbf{Training configuration.} All models use LoRA \citep{hu2022lora} (rank 32, $\alpha=64$), AdamW optimizer, learning rate $10^{-4}$, batch size 32, and one epoch per task. For Moderate-$G_i$, $\rho = 0.2$. Results averaged over three seeds.

\textbf{Baselines.} (1) \textbf{Baseline}: standard fine-tuning; (2) \textbf{Random}: random sampling; (3) \textbf{KL}: KL-divergence regularization against aligned model; (4) \textbf{O-LoRA} \citep{wang2023orthogonal}: orthogonal subspace learning; (5) \textbf{EWC} \citep{kirkpatrick2017overcoming}: Elastic Weight Consolidation, a Fisher-based continual learning method; (6) \textbf{Grad.\ Clip}: gradient norm clipping with clip value 0.5 (best performing among $\{0.1, 0.5, 1.0\}$; see Section~\ref{sec:clipping}).

\textbf{Evaluation.} Task performance via lm-evaluation-harness \citep{eval-harness}. Alignment via: (1) ASR on AdvBench \citep{zou2023universal}, (2) ASR on HarmBench \citep{mazeika2024harmbench}, encompassing direct requests, contextual attacks, and optimization-based jailbreaks, (3) TruthfulQA \citep{lin2022truthfulqa}, (4) commonsense reasoning (ARC-C, BoolQ, HellaSwag, Winogrande). We report checkpoint-averaged values (mean $\pm$ std over three seeds) unless otherwise noted; the high ASR variances in some rows reflect natural variation across the four training stages rather than instability within a single stage.

\subsection{Alignment Preservation}

We first evaluate safety and general capabilities averaged across all training checkpoints and three seeds. Table~\ref{tab:alignment} presents attack success rates (ASR), truthfulness (TruthfulQA), and general capabilities (ARC-C, BoolQ, HellaSwag, Winogrande).

\begin{table}[t]
\centering
\setlength{\tabcolsep}{3pt}
\begin{adjustbox}{width=0.45\textwidth}
\begin{tabular}{llcccc}
\toprule
\textbf{Model} & \textbf{Method} & \textbf{Avg Perf.} & \textbf{BWT}$\uparrow$ & \textbf{FM}$\downarrow$ & \textbf{Max Drop}$\downarrow$ \\
\midrule
\multirow{5}{*}{\rotatebox{90}{\scriptsize Qwen2.5}} 
& Baseline & 59.1 & $-1.7$ & $1.7$ & $11.8$ \\
& Random & 58.9 & $+0.4$ & $-0.4$ & $10.6$ \\
& KL & 61.0 & $-0.9$ & $2.8$ & $5.0$ \\
& O-LoRA & 55.7 & $-8.8$ & $11.4$ & $21.4$ \\
& Moderate-$G_i$ & \textbf{60.9} & $-1.8$ & $1.8$ & $\mathbf{2.7}$ \\
\midrule
\multirow{5}{*}{\rotatebox{90}{\scriptsize LLaMA-3.1}} 
& Baseline & 46.5 & $+0.2$ & $-0.2$ & $8.4$ \\
& Random & 46.8 & $+0.8$ & $-0.8$ & $4.6$ \\
& KL & 46.4 & $-1.1$ & $1.1$ & $8.7$ \\
& O-LoRA & \textbf{47.9} & $\mathbf{+6.4}$ & $\mathbf{-4.5}$ & $-$ \\
& Moderate-$G_i$ & 46.4 & $-1.7$ & $1.7$ & $5.4$ \\
\midrule
\multirow{5}{*}{\rotatebox{90}{\scriptsize Qwen3}} 
& Baseline & 54.2 & $-18.5$ & $18.5$ & $32.5$ \\
& Random & 55.0 & $-19.8$ & $19.8$ & $23.2$ \\
& KL & 54.8 & $-15.5$ & $15.5$ & $26.0$ \\
& O-LoRA & 58.8 & $-12.4$ & $12.4$ & $15.3$ \\
& \textbf{Moderate-$G_i$} & \textbf{58.1} & $\mathbf{-4.3}$ & $\mathbf{4.3}$ & $\mathbf{5.6}$ \\
\bottomrule
\end{tabular}
\end{adjustbox}
\caption{Continual learning performance with forgetting metrics. Avg Perf.: checkpoint-averaged accuracy. BWT: Backward Transfer (higher is better). FM: Forgetting Measure (lower is better). Max Drop: worst single-step performance drop. Moderate-$G_i$ achieves 14.2\% BWT improvement and $5.8\times$ reduction in max drop on Qwen3.}
\label{tab:continual_full}
\end{table}
\paragraph{Safety preservation.} Moderate-$G_i$ achieves substantially lower attack success rates throughout continual learning. On Qwen2.5, our method achieves 10.2\% ASR versus 36.7\% (Baseline), 31.1\% (Random), 33.5\% (KL), and 16.5\% (O-LoRA)—representing 3.6$\times$ reduction over Baseline. EWC achieves 17.4\% ASR but notably \emph{worsens} safety on LLaMA-3.1 (40.2\% vs.\ 44.2\% baseline), demonstrating that Fisher-based regularization, designed to protect task-critical parameters, does not address alignment preservation. Gradient clipping provides only marginal improvement (31.2\% on Qwen2.5), confirming that the issue is sample-specific rather than purely magnitude-based (see Section~\ref{sec:clipping}). On Qwen3, Moderate-$G_i$ (6.0\%) matches O-LoRA (6.8\%), both substantially outperforming Baseline (16.6\%) and Random (11.8\%). On LLaMA-3.1, Moderate-$G_i$ (18.3\%) improves over Baseline (44.2\%) and KL (43.6\%). KL regularization provides minimal benefit despite explicitly constraining distribution drift, while O-LoRA performs well but requires architectural modifications.

These results validate our hypothesis: high-gradient samples reverse safety training during fine-tuning, and filtering them preserves alignment guardrails.

\paragraph{Broader safety evaluation.} To assess generalization beyond AdvBench, we evaluate on HarmBench across the full continual learning pipeline on LLaMA-3.1-8B (Table~\ref{tab:harmbench}). Our method achieves $5.6\times$ lower ASR than baseline on HarmBench, with consistent improvements over all baselines, including EWC, demonstrating that safety gains generalize to more diverse attack vectors.

\begin{table}[t]
\centering\small
\begin{tabular}{lc}
\toprule
\textbf{Method} & \textbf{HarmBench ASR}$\downarrow$ \\
\midrule
Baseline & 27.8 {\scriptsize(3.7)} \\
Random & 17.2 {\scriptsize(5.6)} \\
KL & 27.7 {\scriptsize(1.0)} \\
EWC & 31.0 {\scriptsize(6.7)} \\
O-LoRA & 10.0 {\scriptsize(1.4)} \\
Moderate-$G_i$ & \textbf{5.0} {\scriptsize(3.4)} \\
\bottomrule
\end{tabular}
\caption{HarmBench ASR on LLaMA-3.1-8B across the full continual learning pipeline. Moderate-$G_i$ achieves $5.6\times$ reduction over baseline, generalizing to diverse attack types.}
\label{tab:harmbench}
\end{table}

\paragraph{Cross-architecture variation.} Safety improvements vary across families: Qwen2.5 shows $3.6\times$ ASR reduction, Qwen3 shows $2.8\times$, while LLaMA-3.1 shows $2.4\times$ with no task performance gains over baseline. This variation likely reflects differences in alignment procedures or architectural factors.

\paragraph{Truthfulness and general capabilities.} Moderate-$G_i$ maintains factual accuracy (42.5-42.8\% on TruthfulQA, matching or exceeding baselines) and general capabilities (competitive on ARC-C, BoolQ, HellaSwag, Winogrande, with differences typically within 1-2 points). Sample selection preserves model competencies while improving safety. The complementary strengths of O-LoRA suggest that combining data-centric selection with architectural constraints may yield further improvements.
\begin{figure*}[t]
\centering
\small
\includegraphics[width=\textwidth]{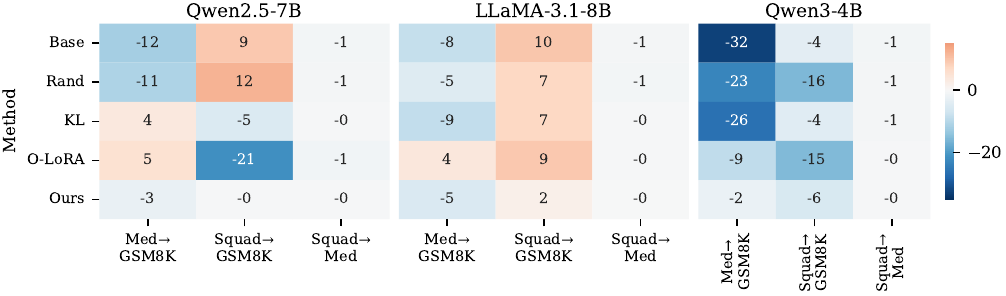}
\caption{Task interference matrix. Negative values (blue) indicate interference; positive (red) indicates transfer. Moderate-$G_i$ consistently minimizes cross-task interference.}
\label{fig:interference}
\end{figure*}

\subsection{Continual Learning Performance}

Table~\ref{tab:continual_performance} presents checkpoint-averaged performance, computed as the mean accuracy across all three evaluation tasks (GSM8K, MedMCQA, Squad\_v2) at each training stage. This metric captures both the model's competence on the current task and its retention of previous capabilities, which is the central challenge in continual learning.

Our gradient-based selection consistently outperforms baselines in maintaining balanced performance across tasks. On Qwen2.5, Moderate-$G_i$ achieves 60.9\% average performance, a 2.0 percentage point improvement over Random (58.9\%) and 1.8 points over Baseline (59.08\%). The advantage grows more pronounced on Qwen3, where our method achieves 58.1\% versus 55.0\% for Random and 54.2\% for Baseline, representing a 3.9 percentage point gain.

\begin{figure}[t]
\centering
\includegraphics[width=0.48\textwidth]{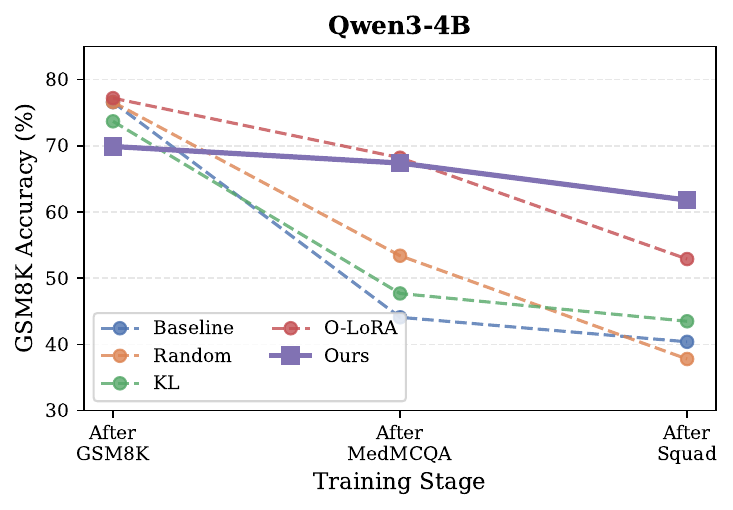}
\caption{GSM8K trajectory on Qwen3. Moderate-$G_i$ maintains gradual degradation while baselines exhibit catastrophic drops.}
\label{fig:forgetting}
\end{figure}
\subsection{Catastrophic Forgetting Analysis}
\label{sec:forgetting}
We analyze catastrophic forgetting using standard continual learning metrics from \citep{lopez2017gradient}: \textit{Backward Transfer} (BWT), measuring how learning new tasks affects previous task performance, and \textit{Forgetting Measure} (FM), quantifying the gap between peak and final accuracy.

Table~\ref{tab:continual_full} extends our performance results with these metrics. On Qwen3, Moderate-$G_i$ achieves BWT of $-4.3\%$ compared to $-18.5\%$ for Baseline—a \textbf{14.2 percentage point improvement}. The forgetting measure confirms this: 4.3\% versus 18.5\%, a $4.3\times$ reduction. These gains are largest on Qwen3, where baselines suffer severe forgetting.
\paragraph{Task Interference Analysis.}
Figure~\ref{fig:interference} reveals \textit{why} methods differ: MedMCQA training causes $-32.5\%$ interference with GSM8K for Baseline on Qwen3, while Moderate-$G_i$ experiences only $-2.5\%$—a \textbf{30 percentage point reduction}. Total task interference drops from $-37.1\%$ to $-8.5\%$ ($4.4\times$ reduction), explaining why gradient-based selection preserves prior knowledge.
\paragraph{Forgetting Trajectories.}
Figure~\ref{fig:forgetting} tracks GSM8K accuracy on Qwen3. Moderate-$G_i$ retains 61.8\% at the final checkpoint (8.1\% drop from peak), while Baseline and Random collapse to 44.1\% and 53.4\% after MedMCQA training. The ``Max Drop'' column in Table~\ref{tab:continual_full} quantifies this: Moderate-$G_i$ limits worst-case drops to 5.6\% versus 32.5\% for Baseline ($5.8\times$ reduction), preventing the catastrophic single-step forgetting that risks crossing safety basin boundaries.

\section{Conclusion}
We presented an empirical investigation of which training samples cause alignment drift during continual fine-tuning. Our findings show that high-gradient samples accelerate reversion toward pretrained distributions, while moderate-gradient samples enable task learning with minimal alignment loss. Our gradient-based sample selection filters high-gradient samples during fine-tuning, achieving strong alignment preservation on Qwen models and modest improvements on LLaMA-3.1, with consistent gains across task orderings, diverse safety benchmarks (AdvBench, HarmBench), and selection ratios. We demonstrate that standard remedies such as gradient clipping and EWC are insufficient, establishing that the issue is sample-specific. While the 51\% computational overhead may limit applicability for very large models, our findings demonstrate that sample selection provides a practical, architecture-agnostic approach to continual safety alignment.
\clearpage






\bibliography{custom}
\appendix
\appendix




\section{Related Work}
\label{sec:related_work}
\paragraph{Alignment Fragility and Theoretical Foundations.}
Safety alignment is surprisingly fragile. Fine-tuning on as few as 10 adversarial examples can compromise safety, while even benign datasets like Alpaca and Dolly inadvertently degrade alignment~\citep{qi2024finetuning,wei2023jailbroken,bach2026curvatureaware}. These vulnerabilities extend to various attack vectors and parameter-efficient methods~\citep{yang2023shadow,lermen2023lora,guan2025benign}. The elasticity framework provides theoretical grounding: language models exhibit an inherent tendency to revert toward pretrained distributions, with elastic force proportional to dataset size~\citep{ji2024language}. This predicts both resistance to initial alignment and rebound toward pretrained behavior under fine-tuning. A complementary geometric perspective conceptualizes alignment through ``safety basins'' in parameter space, where safety exhibits sharp, step-function collapse at basin boundaries~\citep{peng2024navigating}. We adopt the VISAGE metric and extend this framework to understand sample-level contributions to alignment drift.

\paragraph{Continual Learning for LLMs.}
Classical continual learning addresses catastrophic forgetting through parameter regularization~\citep{kirkpatrick2017overcoming} and experience replay~\citep{rolnick2019experience}. LLM-specific studies reveal that forgetting intensifies with model scale~\citep{luo2023empirical} and that vertical continuity (general to specific) differs from horizontal continuity (across time and domains)~\citep{shi2024continual}. Parameter-efficient approaches like O-LoRA learn tasks in orthogonal subspaces to minimize interference~\citep{wang2023orthogonal}. However, these methods focus on task performance rather than alignment preservation, which constitutes a distinct challenge \citep{zheng2025spurious,lianggated}.

\paragraph{Defenses Against Safety Degradation.}
Defenses vary by intervention stage. \emph{Alignment-stage} methods immunize models before fine-tuning through perturbation-aware optimization~\citep{huang2024vaccine}, mapping harmful representations to noise~\citep{rosati2024representation,mukhoti2023finetuning,huang2024lisa,lu2025safe}, or meta-learning to simulate attacks~\citep{tamirisa2024tar,liu2023autodan,zhou2024robust,huang2024booster}. \emph{Fine-tuning-stage} methods constrain adaptation by separating optimization states with proximal regularization~\citep{huang2024lisa} or incorporating safety examples during training. \emph{Post-fine-tuning} methods \citep{peng2025shape,bach2026rethinking} recover safety by projecting weights onto safety-aligned subspaces~\citep{hsu2024safe} or pruning harmful parameters~\citep{huang2024antidote}.

\paragraph{Distinction from One-Shot Safe Fine-Tuning.}
While the above defenses address important challenges, they primarily target single-step fine-tuning scenarios. Continual safety alignment differs in requiring: (1) robustness across sequential adaptations rather than one event, (2) operation without curated safe data at each step, and (3) accounting for cumulative drift across tasks. Our gradient-based selection addresses these issues through a data-centric approach that requires only gradient computation on task data. We compare against KL regularization and O-LoRA as baselines that operate in the continual setting without requiring safe data; a comprehensive comparison against all fine-tuning defenses would require adapting them to sequential multi-task settings.

\section{Extended Background}
\label{app:background}

This appendix provides an extended discussion of the theoretical foundations underlying our work: the safety basin framework and the elasticity phenomenon.

\subsection{Safety Basin Framework: Detailed Treatment}

The safety basin framework, introduced by \citep{peng2024navigating}, provides a geometric perspective on LLM alignment that fundamentally changes how we understand alignment stability. Rather than treating alignment as a binary property or a single metric, this framework reveals that aligned models occupy a connected region in parameter space with distinctive geometric properties.

\paragraph{Safety Landscape Construction.} Given an aligned model with parameters $\theta_{\text{align}}$, the safety landscape is constructed by perturbing along a direction $\mathbf{d}$:
\begin{equation}
    f(\alpha) = S\bigl(\theta_{\text{align}} + \alpha \hat{\mathbf{d}}\bigr)
\end{equation}
where $S(\cdot)$ is a safety metric (e.g., attack success rate on adversarial benchmarks) and $\hat{\mathbf{d}}$ is a normalized perturbation direction. For 2D visualization, two orthogonal directions are used: $f(\alpha, \beta) = S(\theta_{\text{align}} + \alpha \hat{\mathbf{d}}_1 + \beta \hat{\mathbf{d}}_2)$.

\paragraph{Key Empirical Properties.} \citep{peng2024navigating} establish several universal properties across popular open-source LLMs including LLaMA-2, LLaMA-3, Mistral, and Vicuna. First, they observe a \textbf{flat interior}: random perturbations within the basin preserve safety, meaning the aligned model is not a sharp local optimum but sits within a stable region where small random changes to model weights do not immediately compromise safety. Second, there is a \textbf{sharp boundary} where safety exhibits step-function collapse when crossing the basin boundary with minimal graceful degradation. Models are essentially either safe or unsafe, which contrasts sharply with the capability landscape where performance peaks at the origin and gradually declines with perturbation. Third, examining \textbf{fine-tuning trajectories} reveals that fine-tuning on harmful data drags models out of the basin, while fine-tuning on mixed (harmful + safe) data can keep models within the basin, suggesting data composition affects trajectory direction. Finally, the framework reveals significant \textbf{system prompt sensitivity}: removing default system prompts or using roleplaying prompts can reduce VISAGE scores substantially.

\paragraph{VISAGE Metric Details.} The VISAGE (Volumetric Index for Safety Alignment Guided by Explanation) score quantifies basin volume:
\begin{equation}
    \text{VISAGE} = \mathbb{E}_{\alpha \sim U(-a,a)}[S_{\max} - S(\alpha)] \quad \text{s.t.} \quad S < S_{\max}
\end{equation}
where $a$ is the perturbation range and $S_{\max}$ is the maximum safety violation score. In practice, this is computed by sampling $N$ random perturbation directions (we use $N=100$), evaluating safety at multiple perturbation magnitudes for each direction, and computing the average safety margin across all evaluations. Higher VISAGE indicates a larger safety basin and more robust alignment. Empirically, VISAGE scores correlate with model resilience to fine-tuning attacks: models with higher initial VISAGE require more harmful data to break alignment.

\paragraph{Contrast with Capability Landscape.} The safety basin geometry differs fundamentally from the capability landscape. Capability performance peaks at the trained parameters and degrades \emph{gradually} with perturbation magnitude. In contrast, safety is \emph{flat} within the basin and collapses \emph{abruptly} at the boundary. This asymmetry has important implications: while capability loss under fine-tuning is typically gradual and recoverable, safety loss can be sudden and catastrophic.

\subsection{Elasticity Framework: Extended Discussion}

The elasticity framework~\cite{ji2024language} provides theoretical grounding for understanding why alignment is fragile under fine-tuning, drawing on insights from data compression theory.

\paragraph{Compression-Theoretic Foundation.} \citep{ji2024language} model language model training through the lens of data compression. A language model $p_\theta$ trained on dataset $\mathcal{D}$ achieves compression rate $\gamma_{p_\theta}^{\mathcal{D}} = H(\mathcal{D}) / H_{p_\theta}(\mathcal{D})$, where $H(\mathcal{D})$ is the entropy of the data and $H_{p_\theta}(\mathcal{D})$ is the cross-entropy under the model.

\paragraph{Elastic Force Formulation.} The elastic force exerted by dataset $\mathcal{D}_i$ on model parameters is $F_{\text{elastic}} \propto |\mathcal{D}_i| \cdot \Delta D_{\mathrm{KL}}(p_\theta \| p_{\mathcal{D}_i})$. This formulation reveals a critical asymmetry: since pretrain corpora ($|\mathcal{D}_p|$) vastly exceed alignment datasets ($|\mathcal{D}_a|$), the pretrained distribution exerts orders of magnitude stronger ``pull'' on model behavior.

\paragraph{Resistance and Rebound.} The elasticity framework predicts two phenomena. \textbf{Resistance} occurs because pretrained models resist initial alignment due to data volume asymmetry; the alignment process must overcome the elastic force from the massive pretrain corpus. \textbf{Rebound} is the counterintuitive finding that more deeply aligned models revert \emph{faster} to pretrained behavior under perturbation, because deeper alignment represents a larger deviation from the stable pretrained configuration, creating stronger restoring force.

\paragraph{Experimental Validation.} \citep{ji2024language} validate elasticity through several experiments. Training to reverse alignment (moving from aligned to unaligned) consistently shows lower training loss than forward alignment, confirming the asymmetric difficulty. Models trained with more positive (aligned) data initially perform better but deteriorate faster when fine-tuned with negative data. Larger models exhibit stronger rebound effects, with faster initial performance decline and slower subsequent decline.

\paragraph{Connection to Our Work.} We extend the elasticity framework to the sample level. While \citep{ji2024language} characterizes aggregate behavior under fine-tuning, we show that individual samples activate elastic reversion unequally. High-gradient samples, where the aligned model's predictions diverge substantially from fine-tuning targets, represent points of ``alignment tension'' that preferentially activate the reversion force.

\section{Extended Experimental Setup}
\label{app:setup}

\subsection{Model Details}

We conduct experiments on three model families representing different architectures and alignment procedures. \textbf{LLaMA 3.1 8B Instruct}~\cite{llama3} is Meta's instruction-tuned model with safety alignment through RLHF, featuring 32 layers, 32 attention heads, and 4096 hidden dimension. \textbf{Qwen-2.5 7B Instruct}~\cite{qwen2.5} is Alibaba's instruction-tuned model with multi-lingual capabilities, using 28 layers, 28 attention heads, and 3584 hidden dimension. \textbf{Qwen3-4B Instruct}~\cite{yang2025qwen3} is a smaller variant with efficient architecture: 24 layers, 24 attention heads, and 2560 hidden dimension. All models have undergone alignment training and exhibit safety behaviors such as refusing harmful requests and providing helpful responses.

\subsection{Dataset Details}

\paragraph{Dolly.} The Databricks Dolly dataset~\cite{DatabricksBlog2023DollyV2} contains 15,000 instruction-following examples distributed across diverse categories: open QA (2,584 examples), closed QA (1,750), summarization (1,250), information extraction (1,500), creative writing (1,500), classification (2,000), and brainstorming (4,416). Dolly is a benign dataset containing no adversarially designed harmful content, making it suitable for studying unintentional alignment degradation.

\paragraph{Continual Learning Tasks.} We use GSM8K~\cite{cobbe2021gsm8k} containing 8,500 grade-school math problems requiring multi-step reasoning, MedMCQA~\cite{pmlr-v174-pal22a} with medical multiple-choice questions from Indian medical entrance exams, and SQuAD v2~\cite{rajpurkar-etal-2018-know} for reading comprehension, including unanswerable questions.





\subsection{Evaluation Details}
\label{sec:eval_details}

\paragraph{Attack Success Rate (ASR).} 
We evaluate safety using AdvBench~\citep{zou2023universal} with 520 harmful queries. To assess whether model responses are harmful, we employ Llama-Guard-3-8B~\citep{metallamaguard2024} as an automated safety classifier. For each prompt-response pair $(p_i, r_i)$, Llama-Guard evaluates the conversation against multiple safety categories including:
\begin{enumerate}
    \item Illegal activities
    \item Explicit content
    \item Hate speech
    \item Violence
    \item Personal information disclosure
    \item Harassment
    \item Malicious code
    \item Scams or fraud
\end{enumerate}

A response is classified as ``UNSAFE'' if Llama-Guard detects violations in any category. The Attack Success Rate is computed as the proportion of responses classified as unsafe:
{\small
\begin{equation}
    \text{ASR} = \frac{|\{(p_i, r_i) : \textsc{LlamaGuard}(p_i, r_i) = \text{UNSAFE}\}|}{N}
\end{equation}
}
where $N = 520$ is the total number of AdvBench queries.

\paragraph{Task Performance.} Evaluated using lm-evaluation-harness~\cite{eval-harness} with default settings for each benchmark.

\section{Gradient Direction Analysis: Methodology}
\label{appendix:gradient_direction}

This appendix provides methodological details for the gradient direction analysis presented in Section~\ref{sec:direction_analysis}.

\subsection{The High-Dimensionality Challenge}

Computing cosine similarity between gradients and the reversion direction $\mathbf{r} = \theta_{\text{pretrain}} - \theta_{\text{aligned}}$ in billion-dimensional parameter spaces faces concentration of measure: vectors become nearly orthogonal regardless of true underlying alignment. Our preliminary experiments confirmed this issue, with full-parameter cosine similarities of approximately 0.006 even when significant correlations existed.

\subsection{TopK-Cosine Metric}

To address this challenge, we introduce TopK-Cosine, which focuses on the parameters most modified during alignment training. For parameter subset $S$, let $I_k$ index the $k$ dimensions with largest $|r_j|$:
\begin{equation}
    \text{TopK-Cos}(\mathbf{g}, \mathbf{r}; S, k) = \frac{\mathbf{g}_{I_k} \cdot \mathbf{r}_{I_k}}{\|\mathbf{g}_{I_k}\| \|\mathbf{r}_{I_k}\|}
\end{equation}
This metric remains normalized (independent of gradient magnitude) while focusing on alignment-critical parameters. We use $k=1000$ throughout our analysis.

Table~\ref{tab:metric_comparison} validates TopK-Cosine against standard cosine similarity for Qwen2.5-1.5B last-layer V projection. Both metrics yield similar correlation values ($r\approx0.40$--$0.50$), confirming they capture the same underlying effect. However, TopK-Cosine provides approximately $4\times$ higher absolute values (0.119 vs 0.029), improving interpretability.

\begin{table}[h]
\centering
\small
\begin{tabular}{lcccc}
\toprule
\textbf{Metric} & \textbf{HIGH} & \textbf{MOD} & $\mathbf{r}$ & $\mathbf{p}$ \\
\midrule
Full Cosine & 0.029 & 0.027 & 0.50 & $<10^{-3}$ \\
TopK-Cosine & 0.119 & 0.104 & 0.41 & $<10^{-3}$ \\
\bottomrule
\end{tabular}
\caption{Comparison of TopK-Cosine and full cosine for Qwen2.5-1.5B last-layer V projection. Similar correlations confirm consistency; higher TopK-Cosine values improve interpretability.}
\label{tab:metric_comparison}
\end{table}

\subsection{Experimental Protocol}

\paragraph{Reversion Direction.}
For each aligned model, we compute $\mathbf{r} = \theta_{\text{pretrain}} - \theta_{\text{aligned}}$ using matched base models: Qwen2.5-1.5B for Qwen2.5-1.5B-Instruct and Llama-3.1-8B for Llama-3.1-8B-Instruct.

\paragraph{Gradient Collection.}
We compute per-sample gradients $\mathbf{g}_i = \nabla_\theta \mathcal{L}(x_i, y_i; \theta_{\text{aligned}})$ on 500 randomly sampled examples from Dolly, recording gradient norms $G_i = \|\mathbf{g}_i\|_2$.

\paragraph{Parameter Subsets.}
We partition parameters by layer position (last layer, middle third) and component type (Q/K/V/O projections, MLP). This enables localization of directional effects.

\paragraph{Statistical Analysis.}
Samples are stratified into HIGH (top 20\%) and MODERATE (middle 20\%) by gradient norm. We compute group means and Pearson correlation between gradient norm and TopK-Cosine for each parameter subset.

\subsection{Additional Parameter Subsets}

Table~\ref{tab:additional_params} presents TopK-Cosine for additional parameter configurations not shown in the main paper.

\begin{table*}[h]
\centering
\small
\begin{tabular}{llcccc}
\toprule
\textbf{Model} & \textbf{Config} & \textbf{HIGH} & \textbf{MOD} & $\mathbf{r}$ & $\mathbf{p}$ \\
\midrule
\multirow{3}{*}{Qwen2.5-1.5B} 
& last1\_MLP & 0.014 & 0.014 & 0.27 & $<10^{-3}$ \\
& last1\_QKVO & 0.212 & 0.191 & 0.36 & $<10^{-3}$ \\
& last1\_all & 0.044 & 0.043 & 0.34 & $<10^{-3}$ \\
\midrule
\multirow{3}{*}{LLaMA-3.1-8B}
& last1\_O & $-$0.098 & $-$0.092 & 0.09 & 0.22 \\
& last1\_QKVO & $-$0.093 & $-$0.086 & $-$0.07 & 0.31 \\
& last1\_all & $-$0.008 & $-$0.005 & $-$0.05 & 0.50 \\
\bottomrule
\end{tabular}
\caption{TopK-Cosine for additional parameter configurations.}
\label{tab:additional_params}
\end{table*}

For Qwen2.5-1.5B, all final-layer configurations show positive correlation, with V/O projections exhibiting the strongest signal. For LLaMA-3.1-8B, the MLP layer is the primary contributor to directional alignment, while attention projections show weak or no effect.

\subsection{Interpretation}

The localization of directional effects to different components across architectures likely reflects differences in alignment training procedures. Qwen2.5's alignment appears to emphasize modifications to attention output transformations (V/O projections), while LLaMA's alignment concentrates in MLP layers. Despite this variation, the consistent pattern across both models is that directional alignment appears in final-layer parameters and is absent in middle layers, supporting the hypothesis that high-gradient samples reverse alignment-critical modifications.

\section{Additional Experimental Results}
\label{app:additional}
\subsection{What Gets Filtered: Sample Audit}
\label{sec:sample_audit}
To characterize filtered samples and assess potential fairness concerns, we audit 10,000 Dolly samples using Qwen2.5-7B-Instruct, partitioning by gradient norm into three groups.

\begin{table}[t]
\centering\small
\begin{adjustbox}{width=0.48\textwidth}
\begin{tabular}{lccc}
\toprule
\textbf{Property} & \textbf{Low-$G_i$} & \textbf{Moderate-$G_i$} & \textbf{High-$G_i$} \\
\midrule
Mean gradient norm & 1.35 & 3.89 & 16.77 \\
Mean loss & 1.67 & 2.33 & 5.23 \\
Mean answer tokens & 201.7 & 54.6 & 11.5 \\
\bottomrule
\end{tabular}
\end{adjustbox}
\caption{Sample characteristics by gradient group. High-$G_i$ samples have short answers (11.5 tokens) and high per-token loss, indicating format mismatch rather than semantic dissimilarity.}
\label{tab:sample_audit}
\end{table}

High-$G_i$ samples are format mismatches, not content-based outliers. They are dominated by short-answer tasks (classification 28.6\%, closed QA 15.2\%), averaging only 11.5 response tokens with high per-token loss (5.23). Large gradients arise because the aligned model's verbose output distribution diverges from terse targets---not because the content is semantically dissimilar. This indicates no fairness concern: filtering targets output format mismatch, not content or demographics.

\subsection{Task-Specific Performance by Checkpoint}
\begin{table*}[h]
\centering
\begin{adjustbox}{width=\textwidth}
\begin{tabular}{llccccccccc}
\toprule
& & \multicolumn{3}{c}{\textbf{After GSM8K}} & \multicolumn{3}{c}{\textbf{After MedMCQA}} & \multicolumn{3}{c}{\textbf{After SQuAD}} \\
\cmidrule(lr){3-5} \cmidrule(lr){6-8} \cmidrule(lr){9-11}
\textbf{Model} & \textbf{Method} & \textbf{GSM8K} & \textbf{Med} & \textbf{SQuAD} & \textbf{GSM8K} & \textbf{Med} & \textbf{SQuAD} & \textbf{GSM8K} & \textbf{Med} & \textbf{SQuAD} \\
\midrule
\multirow{6}{*}{Qwen2.5-7B} 
& Random & 64.4 {\scriptsize$\pm$ 1.6} & 57.5 {\scriptsize$\pm$ 0.3} & 51.1 {\scriptsize$\pm$ 0.8} & 53.8 {\scriptsize$\pm$ 11.1} & 60.3 {\scriptsize$\pm$ 0.5} & 56.4 {\scriptsize$\pm$ 1.2} & 65.9 {\scriptsize$\pm$ 3.0} & 59.6 {\scriptsize$\pm$ 0.4} & 61.0 {\scriptsize$\pm$ 4.7} \\
& Baseline & 67.1 {\scriptsize$\pm$ 0.4} & 57.1 {\scriptsize$\pm$ 0.3} & 50.3 {\scriptsize$\pm$ 0.1} & 55.3 {\scriptsize$\pm$ 1.3} & 61.1 {\scriptsize$\pm$ 0.0} & 56.8 {\scriptsize$\pm$ 1.1} & 64.5 {\scriptsize$\pm$ 2.3} & 60.2 {\scriptsize$\pm$ 0.1} & 59.3 {\scriptsize$\pm$ 2.4} \\
& KL & 67.1 {\scriptsize$\pm$ 0.4} & 56.8 {\scriptsize$\pm$ 0.1} & 50.5 {\scriptsize$\pm$ 0.2} & 70.8 {\scriptsize$\pm$ 0.6} & 61.0 {\scriptsize$\pm$ 0.2} & 57.4 {\scriptsize$\pm$ 0.2} & 65.8 {\scriptsize$\pm$ 4.6} & 60.5 {\scriptsize$\pm$ 0.2} & 59.0 {\scriptsize$\pm$ 3.3} \\
& O-LoRA & 56.6 {\scriptsize$\pm$ 2.1} & 56.3 {\scriptsize$\pm$ 0.0} & 51.1 {\scriptsize$\pm$ 0.3} & 61.8 {\scriptsize$\pm$ 0.8} & 61.0 {\scriptsize$\pm$ 0.3} & 54.9 {\scriptsize$\pm$ 0.9} & 40.4 {\scriptsize$\pm$ 1.8} & 59.6 {\scriptsize$\pm$ 0.2} & 59.1 {\scriptsize$\pm$ 1.1} \\
\cmidrule{2-11}
& \textbf{Moderate-$G_i$} & 68.6 {\scriptsize$\pm$ 1.8} & 56.7 {\scriptsize$\pm$ 0.5} & 51.7 {\scriptsize$\pm$ 0.7} & 65.9 {\scriptsize$\pm$ 1.6} & 58.7 {\scriptsize$\pm$ 0.2} & 57.3 {\scriptsize$\pm$ 1.4} & 65.4 {\scriptsize$\pm$ 4.0} & 58.2 {\scriptsize$\pm$ 0.5} & 65.4 {\scriptsize$\pm$ 5.6} \\
\midrule
\multirow{6}{*}{LLaMA-3.1-8B}
& Random & 19.8 {\scriptsize$\pm$ 4.3} & 58.7 {\scriptsize$\pm$ 0.4} & 54.6 {\scriptsize$\pm$ 0.9} & 15.2 {\scriptsize$\pm$ 0.9} & 60.0 {\scriptsize$\pm$ 0.3} & 57.5 {\scriptsize$\pm$ 2.1} & 22.4 {\scriptsize$\pm$ 6.8} & 59.1 {\scriptsize$\pm$ 0.4} & 74.2 {\scriptsize$\pm$ 1.8} \\
& Baseline & 23.4 {\scriptsize$\pm$ 1.2} & 59.0 {\scriptsize$\pm$ 0.2} & 50.7 {\scriptsize$\pm$ 0.3} & 15.0 {\scriptsize$\pm$ 2.0} & 60.1 {\scriptsize$\pm$ 0.4} & 56.2 {\scriptsize$\pm$ 0.9} & 24.6 {\scriptsize$\pm$ 1.7} & 59.4 {\scriptsize$\pm$ 0.4} & 70.4 {\scriptsize$\pm$ 2.4} \\
& KL & 23.8 {\scriptsize$\pm$ 1.8} & 58.8 {\scriptsize$\pm$ 0.2} & 51.0 {\scriptsize$\pm$ 0.2} & 15.1 {\scriptsize$\pm$ 1.5} & 59.7 {\scriptsize$\pm$ 0.3} & 56.8 {\scriptsize$\pm$ 0.5} & 22.1 {\scriptsize$\pm$ 4.8} & 59.2 {\scriptsize$\pm$ 0.2} & 70.7 {\scriptsize$\pm$ 2.1} \\
& O-LoRA & 18.8 {\scriptsize$\pm$ 2.2} & 59.3 {\scriptsize$\pm$ 0.2} & 52.6 {\scriptsize$\pm$ 0.5} & 22.7 {\scriptsize$\pm$ 1.3} & 59.7 {\scriptsize$\pm$ 0.1} & 54.6 {\scriptsize$\pm$ 0.7} & 32.0 {\scriptsize$\pm$ 2.1} & 59.4 {\scriptsize$\pm$ 0.2} & 72.0 {\scriptsize$\pm$ 0.3} \\
\cmidrule{2-11}
& \textbf{Moderate-$G_i$} & 24.0 {\scriptsize$\pm$ 4.4} & 59.1 {\scriptsize$\pm$ 0.2} & 52.5 {\scriptsize$\pm$ 1.3} & 18.6 {\scriptsize$\pm$ 3.8} & 58.3 {\scriptsize$\pm$ 2.1} & 56.3 {\scriptsize$\pm$ 1.5} & 20.7 {\scriptsize$\pm$ 3.1} & 58.1 {\scriptsize$\pm$ 1.8} & 70.2 {\scriptsize$\pm$ 3.8} \\
\midrule
\multirow{6}{*}{Qwen3-4B}
& Random & 76.6 {\scriptsize$\pm$ 2.6} & 55.2 {\scriptsize$\pm$ 0.5} & 50.1 {\scriptsize$\pm$ 0.0} & 53.4 {\scriptsize$\pm$ 8.3} & 58.3 {\scriptsize$\pm$ 0.2} & 50.4 {\scriptsize$\pm$ 0.5} & 37.8 {\scriptsize$\pm$ 7.3} & 57.5 {\scriptsize$\pm$ 0.2} & 55.4 {\scriptsize$\pm$ 4.6} \\
& Baseline & 76.6 {\scriptsize$\pm$ 0.6} & 55.5 {\scriptsize$\pm$ 0.2} & 50.1 {\scriptsize$\pm$ 0.0} & 44.1 {\scriptsize$\pm$ 2.3} & 59.4 {\scriptsize$\pm$ 0.4} & 50.1 {\scriptsize$\pm$ 0.0} & 40.4 {\scriptsize$\pm$ 5.0} & 58.5 {\scriptsize$\pm$ 0.0} & 52.9 {\scriptsize$\pm$ 1.9} \\
& KL & 73.7 {\scriptsize$\pm$ 2.7} & 55.5 {\scriptsize$\pm$ 0.0} & 50.1 {\scriptsize$\pm$ 0.0} & 47.7 {\scriptsize$\pm$ 5.0} & 59.3 {\scriptsize$\pm$ 0.4} & 50.2 {\scriptsize$\pm$ 0.2} & 43.5 {\scriptsize$\pm$ 4.3} & 58.5 {\scriptsize$\pm$ 0.7} & 54.6 {\scriptsize$\pm$ 6.9} \\
& O-LoRA & 77.2 {\scriptsize$\pm$ 0.8} & 55.2 {\scriptsize$\pm$ 0.1} & 50.1 {\scriptsize$\pm$ 0.0} & 68.2 {\scriptsize$\pm$ 3.2} & 59.1 {\scriptsize$\pm$ 0.3} & 50.1 {\scriptsize$\pm$ 0.0} & 52.9 {\scriptsize$\pm$ 1.6} & 58.6 {\scriptsize$\pm$ 0.1} & 58.2 {\scriptsize$\pm$ 0.5} \\
\cmidrule{2-11}
& \textbf{Moderate-$G_i$} & 69.9 {\scriptsize$\pm$ 1.0} & 54.7 {\scriptsize$\pm$ 0.9} & 50.1 {\scriptsize$\pm$ 0.0} & 67.4 {\scriptsize$\pm$ 9.8} & 55.1 {\scriptsize$\pm$ 0.3} & 50.2 {\scriptsize$\pm$ 0.3} & 61.8 {\scriptsize$\pm$ 8.7} & 54.7 {\scriptsize$\pm$ 0.3} & 59.0 {\scriptsize$\pm$ 5.7} \\
\bottomrule
\end{tabular}
\end{adjustbox}
\caption{Task-specific performance across continual learning checkpoints for all methods. Each checkpoint evaluates on all three tasks (GSM8K, MedMCQA, Squad v2). Values show mean $\pm$ standard deviation across 3 seeds.}
\label{tab:full_results}
\end{table*}

\paragraph{Overall trends.} Moderate-$G_i$ consistently mitigates catastrophic forgetting on GSM8K across continual learning stages. On Qwen3-4B, our method retains 67.4\% after MedMCQA and 61.8\% after SQuAD v2, compared to baselines' 44-53\% and 38-40\% respectively. MedMCQA training causes the most severe interference with GSM8K across all models, while Squad training shows variable effects, which sometimes recover performance (Qwen2.5-7B Random: 53.78 $\rightarrow$ 65.9\%) and sometimes continue degradation (Qwen3-4B: 53 $\rightarrow$ 38\%).

\paragraph{Training stability.} Moderate-$G_i$ substantially reduces variance compared to Random sampling. The most striking example occurs on Qwen2.5-7B after MedMCQA: Random exhibits 11.1\% standard deviation on GSM8K due to one seed collapsing to 40.9\%, while Moderate-$G_i$ maintains 1.3\% std with no catastrophic failures. This stability advantage persists across models and checkpoints, demonstrating that gradient-based selection produces more reliable continual learning dynamics.

\subsection{Extended Continual Learning Analysis}
\label{app:extended_cl}

This section provides supplementary visualizations for the continual learning analysis in Section~\ref{sec:forgetting}.

\begin{figure*}[t]
\centering
\includegraphics[width=0.8\textwidth]{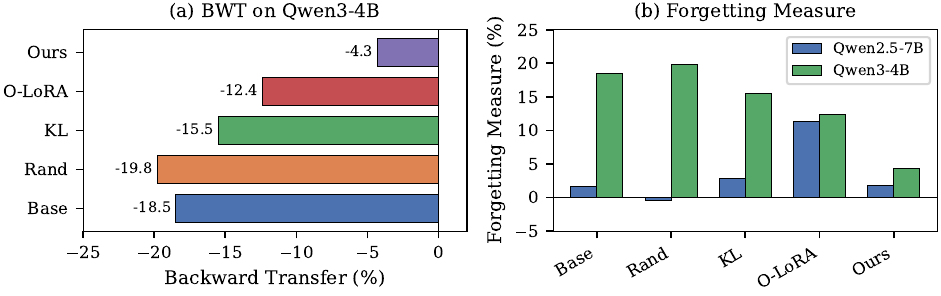}
\caption{Continual learning metrics visualization. (a) BWT on Qwen3-4B shows 14.2\% improvement for Moderate-$G_i$. (b) FM comparison reveals gradient-based selection provides the largest benefits on forgetting-prone models.}
\label{fig:bwt_fm}
\end{figure*}

\paragraph{Backward Transfer and Forgetting Measure Visualization.}
Figure~\ref{fig:bwt_fm} visualizes the BWT and FM metrics from Table~\ref{tab:continual_full}. Panel (a) shows Backward Transfer on Qwen3-4B, where Moderate-$G_i$'s advantage ($-4.3\%$ vs $-18.5\%$ baseline) is visually striking. Panel (b) reveals an important pattern: while methods show similar forgetting on Qwen2.5-7B (FM $\leq 3\%$), they diverge dramatically on Qwen3-4B (4.3\% vs 18-20\%). This suggests gradient-based selection becomes increasingly valuable as models become more susceptible to forgetting.

\begin{figure*}[t]
\centering\small
\includegraphics[width=0.6\textwidth]{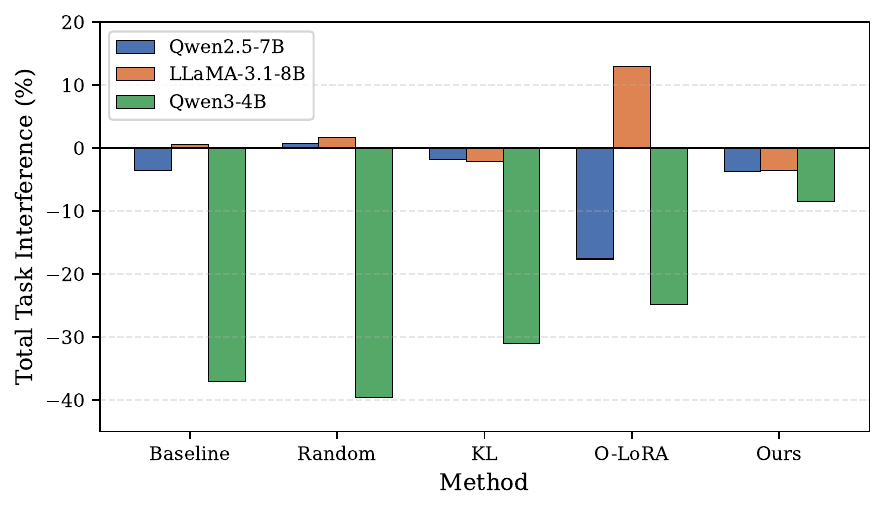}
\caption{Total task interference (sum of all pairwise effects). Moderate-$G_i$ achieves $4.4\times$ reduction on Qwen3-4B.}
\label{fig:total_interference}
\end{figure*}

\paragraph{Total Task Interference.}
Figure~\ref{fig:total_interference} presents total task interference, the sum of all pairwise interference effects from Figure~\ref{fig:interference}. On Qwen3-4B, Baseline accumulates $-37.1\%$ total interference compared to $-8.5\%$ for Moderate-$G_i$ ($4.4\times$ reduction). This cumulative view explains the performance gaps in Table~\ref{tab:continual_full}: each training stage introduces less destructive interference with gradient-based selection, allowing knowledge to accumulate rather than cancel.

\paragraph{Connection to Safety Alignment.}
The metrics reinforce our central thesis: high-gradient samples cause both alignment drift and catastrophic forgetting through the same mechanism:
\begin{enumerate}
    \item Large parameter updates reverse task-specific representations (causing forgetting)
    \item Updates push parameters toward pretrained distributions (elastic reversion)  
    \item Extreme updates risk crossing safety basin boundaries (alignment collapse)
\end{enumerate}
Moderate-$G_i$ addresses all three by preventing the high-gradient samples that trigger them. Consistent improvements across all metrics, especially where safety gains are largest, confirm that gradient-based selection taps into fundamental continual learning dynamics.

\subsection{Why Not Gradient Clipping?}
\label{sec:clipping}
A natural alternative to filtering high-gradient samples is gradient norm clipping, which bounds update magnitude without discarding data. We compare the two approaches on Dolly with Qwen2.5-7B (Table~\ref{tab:clipping}).

\begin{table}[t]
\centering\small
\begin{tabular}{lccc}
\toprule
\textbf{Method} & \textbf{ASR}$\downarrow$ & \textbf{TruthfulQA} & \textbf{Task Avg} \\
\midrule
Baseline (clip=1.0) & 36.7 & 38.2 & 59.1 \\
Clip 0.5 & 31.2 & 38.2 & 59.4 \\
Clip 0.1 & 32.4 & 37.3 & 58.9 \\
Moderate-$G_i$ & \textbf{10.2} & \textbf{42.5} & \textbf{60.9} \\
\bottomrule
\end{tabular}
\caption{Gradient clipping vs.\ sample selection on Dolly with Qwen2.5-7B. Clipping provides only marginal ASR improvement (best: 31.2\% at clip=0.5), while our method achieves $3\times$ further reduction.}
\label{tab:clipping}
\end{table}

Gradient clipping provides only marginal improvement (best: 31.2\% ASR at clip=0.5), while our method achieves 10.2\%---a $3\times$ further reduction. Clipping also fails to improve TruthfulQA (${\sim}38\%$ vs.\ our 42.5\%). This reveals a key distinction: the problem is not gradient magnitude per se, but \emph{which samples} generate those gradients. Clipping attenuates step size but still trains on high-gradient samples---the model still receives a learning signal pushing toward pretrained distributions. Our method removes these samples entirely, preventing their alignment-reversing content from influencing training.

\subsection{Task Order Robustness}
\label{sec:task_order}
To evaluate sensitivity to task ordering, we test two additional orderings and a domain substitution on LLaMA-3.1-8B (Table~\ref{tab:task_order}).

\begin{table}[t]
\centering\small
\begin{tabular}{lcc}
\toprule
\textbf{Method} & \textbf{ASR}$\downarrow$ & \textbf{Task Avg} \\
\midrule
\multicolumn{3}{l}{\textit{Order 1: Dolly $\rightarrow$ MedMCQA $\rightarrow$ GSM8K $\rightarrow$ Squad}} \\
Random & 7.8 & 45.9 \\
EWC & 15.5 & 47.3 \\
KL & 11.0 & 48.2 \\
O-LoRA & 1.2 & 48.4 \\
Moderate-$G_i$ & \textbf{1.0} & \textbf{48.7} \\
\midrule
\multicolumn{3}{l}{\textit{Order 2: Dolly $\rightarrow$ Squad $\rightarrow$ MedMCQA $\rightarrow$ GSM8K}} \\
Random & 15.5 & 51.6 \\
EWC & 12.8 & 50.7 \\
O-LoRA & 4.0 & 49.2 \\
Moderate-$G_i$ & \textbf{0.8} & \textbf{52.3} \\
\midrule
\multicolumn{3}{l}{\textit{Domain substitution (Alpaca replacing Dolly), Qwen2.5-7B}} \\
Baseline & 39.8 & 57.3 \\
Random & 19.0 & 59.1 \\
Moderate-$G_i$ & \textbf{10.4} & \textbf{59.6} \\
\bottomrule
\end{tabular}
\caption{Task order robustness. Our method achieves best or near-best ASR and task accuracy across all configurations, demonstrating Pareto dominance independent of task ordering.}
\label{tab:task_order}
\end{table}

Under alternative orderings, our method achieves both the best safety (ASR 1.0\% and 0.8\%) and the best task performance (48.7\% and 52.3\%) on LLaMA-3.1-8B, demonstrating Pareto dominance. The cross-architecture variation observed in the original ordering likely reflects differences in how model families were aligned during post-training, rather than a fundamental limitation of our approach. Results also generalize across initial datasets (Dolly $\rightarrow$ Alpaca).

\subsection{Computational Overhead}

\begin{table}[h]
\centering
\small
\begin{tabular}{lcc}
\toprule
\textbf{Method} & \textbf{Time/Epoch} & \textbf{Relative Cost} \\
\midrule
Baseline & 14.4 min & 1.00$\times$ \\
KL Regularization & 20.4 min & 1.42$\times$ \\
O-LoRA & 14.8 min & 1.03$\times$ \\
Moderate-$G_i$ (ours) & 21.6 min & 1.5$\times$ \\
\bottomrule
\end{tabular}
\caption{Wall-clock training time on Qwen2.5-7B on Dolly with single H100 GPU.}
\label{tab:overhead}
\end{table}

Our method adds 51\% computational overhead due to gradient computation for candidate filtering. However, this overhead occurs only during training; inference cost is unchanged. The overhead can be reduced through gradient checkpointing, computing gradients only for LoRA parameters, and caching gradient statistics across epochs.

\section{Limitations and Future Work}
\label{app:limitations}

\paragraph{Computational Overhead.} Our method requires additional gradient computation during training (1.5$\times$ baseline cost). While acceptable for most deployment scenarios, this overhead may be prohibitive for very large models or extremely limited compute budgets.

\paragraph{Gradient Approximations.} We compute exact per-sample gradients, which requires sequential backward passes. Gradient approximation techniques such as influence functions or gradient sketching could reduce this cost but may affect selection quality.

\paragraph{Dynamic Selection Threshold.} We use a fixed selection ratio $\rho = 0.2$ throughout training. Adaptive strategies that adjust selection strictness based on observed alignment drift could improve efficiency.

\paragraph{Theoretical Characterization.} While we provide empirical evidence linking gradient magnitude to alignment drift, a formal theoretical analysis connecting sample gradients to safety basin geometry remains future work.

\paragraph{Multi-Modal Models.} Our experiments focus on text-only models. Extending gradient-based selection to vision-language models or other modalities requires further investigation.
\end{document}